# A Digital Simulation Toolkit for Physics-Based Generation of Realistic Experimental Scanning Tunneling Microscopy Images

Huanhuan Zhao[1*], Laxmi Bhurtel[2,3], Connor Vernachio[2,3], Fahmy Paiziah[4], Wonhee Ko[2,3*], Arpan Biswas[3,5*]

[1]Bredesen Center for Interdisciplinary Research, University of Tennessee, Knoxville, USA, 37996

[2]Department of Physics and Astronomy, University of Tennessee, Knoxville, Tennessee 37996, USA

[3]Center for Advanced Materials and Manufacturing, University of Tennessee, 2641 Osprey Vista Way, Knoxville, Tennessee 37920, USA

[4]Department of Computer Science, Hunter College, City University of New York, 695 Park Avenue, New York, New York 10065, USA

[5]University of Tennessee-Oak Ridge Innovation Institute, University of Tennessee, Knoxville, USA, 37996

**Email**: hzhao31@vols.utk.edu, wko@utk.edu, abiswas5@utk.edu

## Abstract

Scanning Tunneling Microscopy (STM) is a widely used tool for characterizing surfaces of materials at the atomic scale, playing a crucial role in discoveries across condensed matter physics and materials science. Despite its extreme spatial resolution, STM is one of the most sensitive microscopy techniques and is highly prone to noise. While existing unsupervised denoising methods such as N2V and N2V2 models are very cheap to train, these are primarily focused on removing the noise and artifacts with minimal recovery of subtle material surface details within artifacts, containing key physical information for material discovery. On the other hand, while supervised methods such as Nonlinear Activation Free Network (NAFNet) and Restormer can offer superior performance than unsupervised approaches, the major bottleneck is the high complexity of the appropriate training data generation. For example, in application of such methods in cleaning STM images, a large amount of paired clean-noisy images is required which are impractical to obtain. To address this global data availability challenge in the field of microscopy, we developed a low-cost physics-driven digital toolkit to rapidly generate large volume of realistic STM images. Firstly, we simulate clean images from a chosen material system. Then, with prior knowledge of the physical characteristics of the artifacts and noise present in STM experiments, we formulate several artifact-noise functions such as Gaussian electronic noise, 1/f flicker noise, scan-line noise, background tilt and mechanical drift. These physically informed noise and artifact components are then added to the simulated clean images to generate realistic STM images. We demonstrated the capability of the proposed digital toolkit for STM image generation into training NAFNet and Restormer models, against unsupervised methods, for denoising synthetic and experimental images of the (111) surfaces of copper and lead, while preserving the physical features such as atoms, defects, and electron waves. To further understand the impact of the developed digital toolkit, we validated on the quality of the downstream image analysis of learning electron wave patterns induced by quantum interference from experimental Cu(111) images. Results show that the supervised models trained on digitally generated AI-ready data can more effectively denoise and learn electron wave patterns on experimental Cu(111) images than benchmarked unsupervised approaches, indicating that the proposed toolkit adequately represents the noise characteristics present in STM measurements and facilitate scientific discovery. The proposed work is readily applicable to other materials and STM imaging conditions, providing a generalizable and extensible tool for generating realistic AI-ready microscopy data to support the development of a broad range of data-driven STM analysis and automation tasks.

## 1. Introduction

Scanning tunneling microscopy (STM) provides atomic level images of material surfaces, revealing electronic structure and discovering key material properties. Thus, over the years, STM has been extensively used for materials characterization[1–6]. STM relies on the tunneling current between the tip and the sample for imaging, which decays exponentially with tip-sample distance. This extreme distance sensitivity grants STM its atomic-scale resolution, but also makes the measurement highly susceptible to noise. Any physical or electronic disturbance affecting the tunneling current can introduce significant artifacts into the image. Noise in STM images can arise from multiple sources, including mechanical vibration, thermal drift, electronic noise, piezo creep and hysteresis, and feedback loop instability[7]. While it is an extremely tedious task to optimize the experimental settings to minimize these artifacts and noise, ignoring them can corrupt the real physical features completely. Because of the complexity of noise in STM images, simple traditional denoising algorithms such as gaussian filtering, band-pass filtering, and wavelet denoising may not recover the true physical features from such corrupted images. These denoising methods tend to blur or over smooth the true physical features in microscopy images.

On the other hand, deep learning models have proven highly effective in dealing with image denoising and restorations. However, the major bottlenecks in model training for STM image cleaning are 1) the scarcity of the microscopy images due to expensive data acquisition and 2) non-availability of clean microscopy images to establish data labelling. Due to the lack of sufficient training STM data, unsupervised deep learning models such as N2V and N2V2 have been developed for microscopy image denoising[8,9]. However, the performance of these unsupervised models remains limited, and they often do not provide a balanced recovery of fine physical features and removal of noise. In other words, significant loss of information from subtle physical features and addition of new artifacts potentially occurs with the removal of the artifact-noise in STM[10]. In our previous work, we developed a global denoising model (GDM) to clean different microscopy images with better trade-offs between feature preservations and noise removal[11]. However, the model requires manual training image pre-processing steps to attain appropriate image cleaning.

Supervised deep learning models, such as NAFNet[12] and Restormer[13], have demonstrated remarkable image restoration performance, achieving state-of-the-art denoising results on natural image benchmarks. However, these methods require large, paired datasets of noisy and clean images for training. For STM, obtaining a perfectly noise-free image is fundamentally impractical as every measurement is subject to unavoidable noise from various external sources. This lack of paired training data represents a critical barrier in applying advanced supervised deep learning to STM denoising. Over the years, researchers have explored both traditional image processing and ML methods for STM image denoising. Van Kempen et al. used a Wiener filter to remove the flicker noise[14]. Fogarty et al. applied linear-regression fitting to remove external noise caused by vibrational or acoustic interference[15]. Fan et al. used a generative network to synthesize periodic noise and construct paired noisy-clean datasets and train a CNN model to remove periodic noise in STM images[16]. Kolev et al. employed the physics-informed synthetic data method to train a diffusion model to correct the tip degradation and achieves super-resolution[17].

However, few attempts have been made to learn different artifacts in STM images, superimposed over one another. Aguilar et al. characterized flicker noise in STM measurements, demonstrating the complexity of experimental noise[18]. Joucken et al. later demonstrated the potential of simulation-based data generation for STM image denoising by adding synthetic noise to simulated graphene images to construct paired clean–noisy datasets for supervised U-Net training[19]. However, their approach was limited to three predefined noise sources—Gaussian noise, row-by-row drift, and a paraboloidal background artifact—and therefore does not capture the broader range of noise and imaging artifacts encountered in STM measurements. Despite these efforts, a general and extensible framework lacks simulating realistic STM images across diverse noise mechanisms, material systems, and imaging conditions. This gap limits the systematic generation of AI-ready microscopy training datasets for ML models, particularly when experimental datasets are scarce or difficult to acquire. In other words, incomplete or inaccurate representations of real microscopy images can lead to inadequate learning by ML models, thereby compromising the effectiveness of STM image denoising and other downstream image analysis tasks. Therefore, there is a need for a general STM image simulation toolkit that integrates multiple experimentally informed noise and artifact models to generate realistic STM datasets that can support the training and evaluation of diverse ML models and downstream image analysis methods across different material systems.

To address this global challenge in data acquisition cost, complexity and availability in the field of STM, we present a low-cost **physics-driven noise-simulation digital toolkit** to rapidly generate large volume of realistic STM images. The contribution of this work is the development of a noise simulation model for STM images which includes five dominant noise sources: **Gaussian noise, row-by-row drift, paraboloid background artifact, flicker noise, and periodic stripe lines.** The proposed simulation model has user-controlled intensity and the density of the individual noise functions, to allow variability in the simulated realistic STM images.

Here, we explain how each type of noise is defined and produced. The noise types included in the toolkit were selected through a combination of literature review and visual inspection of experimental STM images. Gaussian noise, row-by-row drift, and paraboloid background artifacts are common types of noise in STM images, and are used in the STM image noise simulation by Joucken et al.[19] In this work, **Gaussian noise** is randomly sampled from a normal distribution with a mean of zero and standard deviation ranging from 0.01 to 0.03. **Row-by-row drift** is caused by mechanical drift of the STM tip relative to the sample surface during image acquisition. During noise simulation, each row has a drift of up to three pixels compared to the previous row. The **paraboloid background artifact** is a slowly varying curved background commonly observed in scanning probe microscopy (SPM) techniques such as STM and Atomic Force Microscopy (AFM). Different shapes of paraboloid background, including bowl, dome and saddle, are randomly sampled and added to each STM image. We additionally integrate two more types of noise: flicker noise and periodic stripe lines. **Flicker noise** is a $\frac{1}{f^{\beta}}$ like noise that exists in STM and other electronic devices. Aguilar et al. developed a method to characterize the value of β and suggested that its value falls within range (1, 3)[18]. Thus, in our code we randomly sample β's value from (1,3). We first generate all possible frequencies in a scan using FFT and random sample the corresponding amplitude and phase from a normal distribution (with mean of zero, standard deviation of 1). The inverse real FFT then produces a time-domain noise trace. A sequence contains $2N^2$ points were generated for an $N \times N$ image and only the forward segment is retained and written into the image array, while the back-scan segment is discarded, adhering to the real STM scanning process.

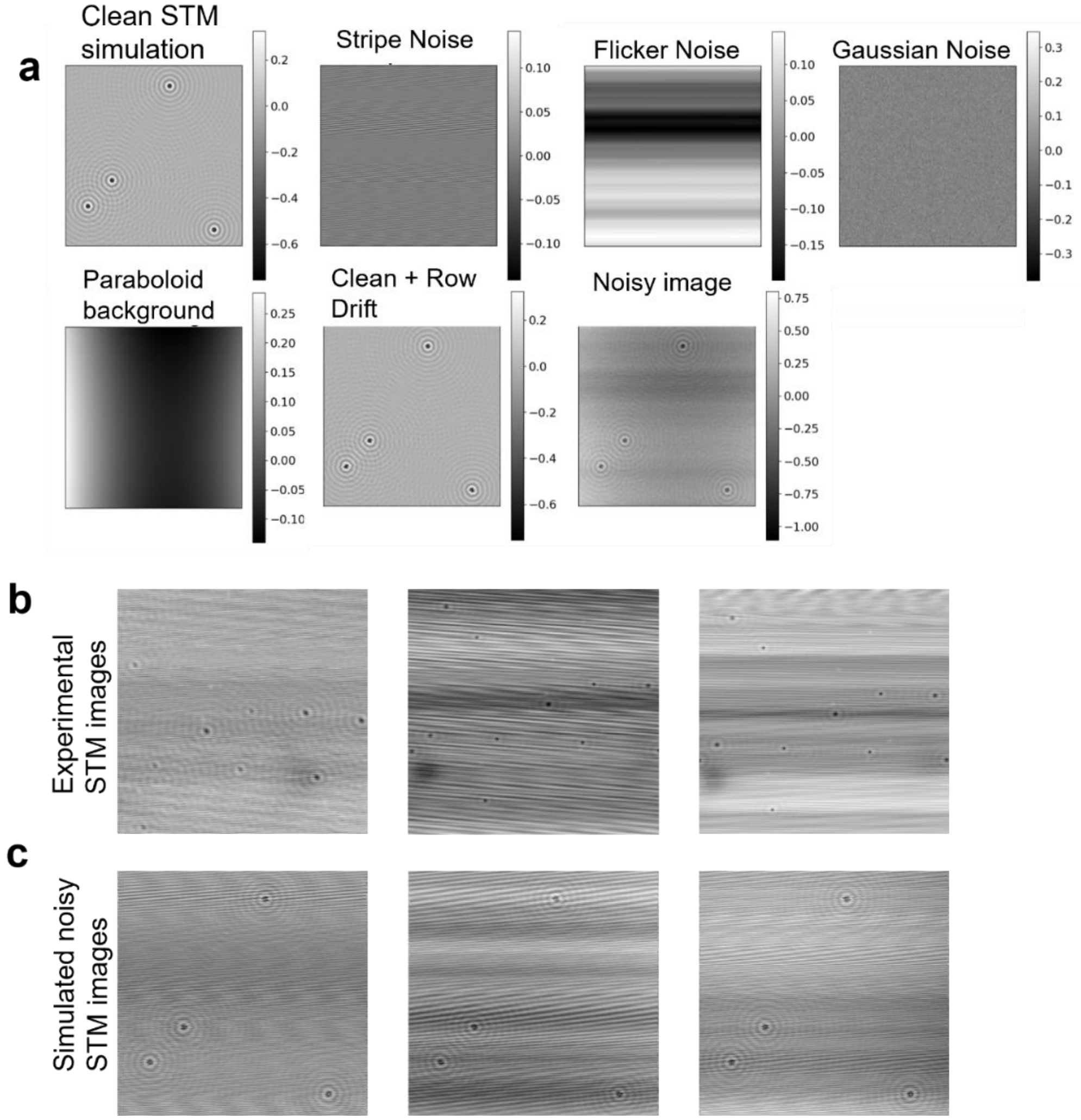

**Figure 1. Noisy STM image simulation**. (a) Illustration of the noise generation pipeline applied to a simulated Cu(111) clean STM image. First row from left to right: clean simulated image, stripe noise, $\frac{1}{f^{\beta}}$ flicker noise, Gaussian noise. Second row from left to right: paraboloid background artifact, combined noisy image, clean image with only row drift, and final noisy image with row drift. (b) show examples of experimental noisy Cu(111) STM images and 1(c) show simulated noisy Cu(111) STM images.

**Stripe noise** is caused by external interruptions such as mechanical vibrations or electronic resonances coupled into the STM. It appears as distinct bands with periodic repetition along the scan direction, and its transitions are sharper and more abrupt compared to flicker noise. For each image, we generate one to five stripes with different frequencies and varying brightness to mimic the experimental noise. All noise components and clean images are first normalized to the range [0, 1]. Each noise type is then randomly assigned a weight ranging from 0 to 0.1, and all weighted components are combined and added to the clean simulated image. Together, these components capture the dominant random and structured artifacts that degrade STM image quality. The overall noise simulation process is illustrated in **Fig.1(a)**, which shows a clean simulated STM image, the first four modeled noise types, the image corrupted by row drift only, and the final noisy image obtained by combining all noise components. Each noise component is generated independently and combined to produce a realistic synthetic noisy STM image. **Fig.1(b)** shows representative experimental STM images of Cu(111) acquired under typical laboratory conditions, showing characteristic noise artifacts including flicker noise, stripe noise, and drift. **Fig.1(c)** shows representative simulated noisy STM images generated by the proposed noise model, demonstrating visual similarity to the experimental images in **Fig.1(b)**.

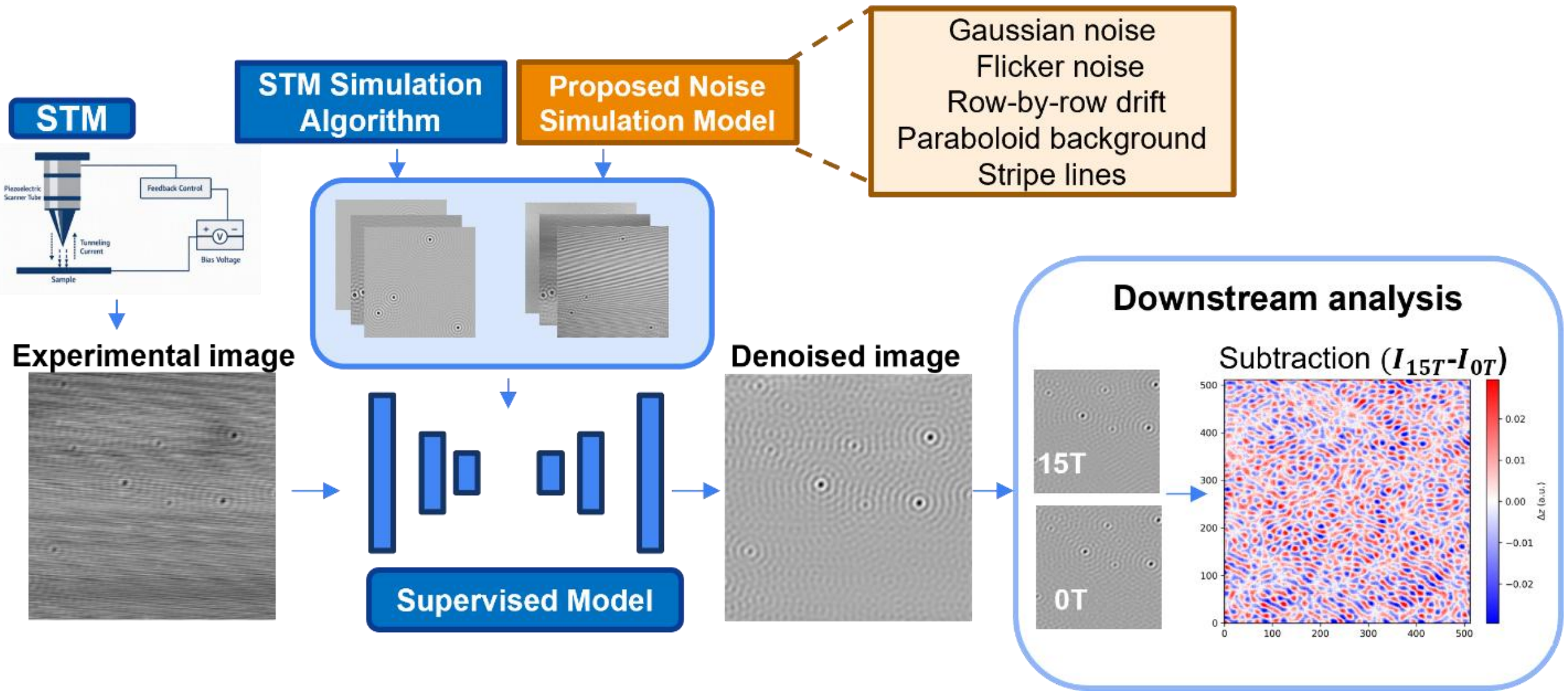

**Figure 2.** Method pipeline illustrating how the proposed physics-driven noise-simulation digital toolkit enables supervised deep learning-based denoising and downstream scientific discovery. In this example, we have demonstrated Cu(111) images, supervised image denoising models and downstream analysis of electron wave patterns with the changes of magnetic field.

To demonstrate the significance of the proposed digital toolkit for rapid generation of AI-ready STM images for supervised image denoising and downstream scientific discovery, we have designed the pipeline as per **Fig. 2.** We first simulated clean STM images using established simulation tools. In the pipeline, we considered the Cu(111) surfaces simulation developed by Fiete, G. A., and Heller, E. J.[20]. Then we added the defined noises to each of the simulated clean images to construct the relevant noisy images, thus generating the paired clean-noisy images. Next, we trained a state-of-the-art image denoising model, such as NAFNet or Restormer, using the generated AI-ready data from the proposed digital toolkit. Then, we apply the experimental STM images over the trained supervised model for denoising and further downstream image analysis for domain specific knowledge extraction.

## 2. Results and Discussion

In this section, we have demonstrated the application of the proposed noise-simulation digital toolkit by training two supervised deep learning models for STM image denoising. Furthermore, we demonstrated the potential of the noise simulation toolkit in an improved downstream image analysis to facilitate scientific discovery.

### 2.1. Denoising of simulated noisy STM images:

After generating simulated paired clean-noisy AI-ready datasets from the proposed digital framework, we trained two state-of-the-art denoising model NAFNet and Restormer. We first evaluated the trained supervised models on simulated noisy images and assessed their denoising performance by comparing with the respective cleaned ground truth images. We further compared the results of benchmarked supervised models such as NAFNet and Restormer with benchmarked unsupervised models such as N2V and GDM. Here, GDM and N2V were trained with one clean simulated Cu(111) image and one experimental Cu(111) image. The evaluation metric used are PSNR and SSIM to compare the performance among these methods.

**Table 1.** Quantitative denoising performance evaluation on simulated Cu(111) STM test images.

| | GDM | N2V | NAFNet | Restormer |
|---|---|---|---|---|
| **PSNR** | **22.64** | **22.81** | **26.60** | **36.33** |
| **SSIM** | **0.79** | **0.58** | **0.94** | **0.98** |

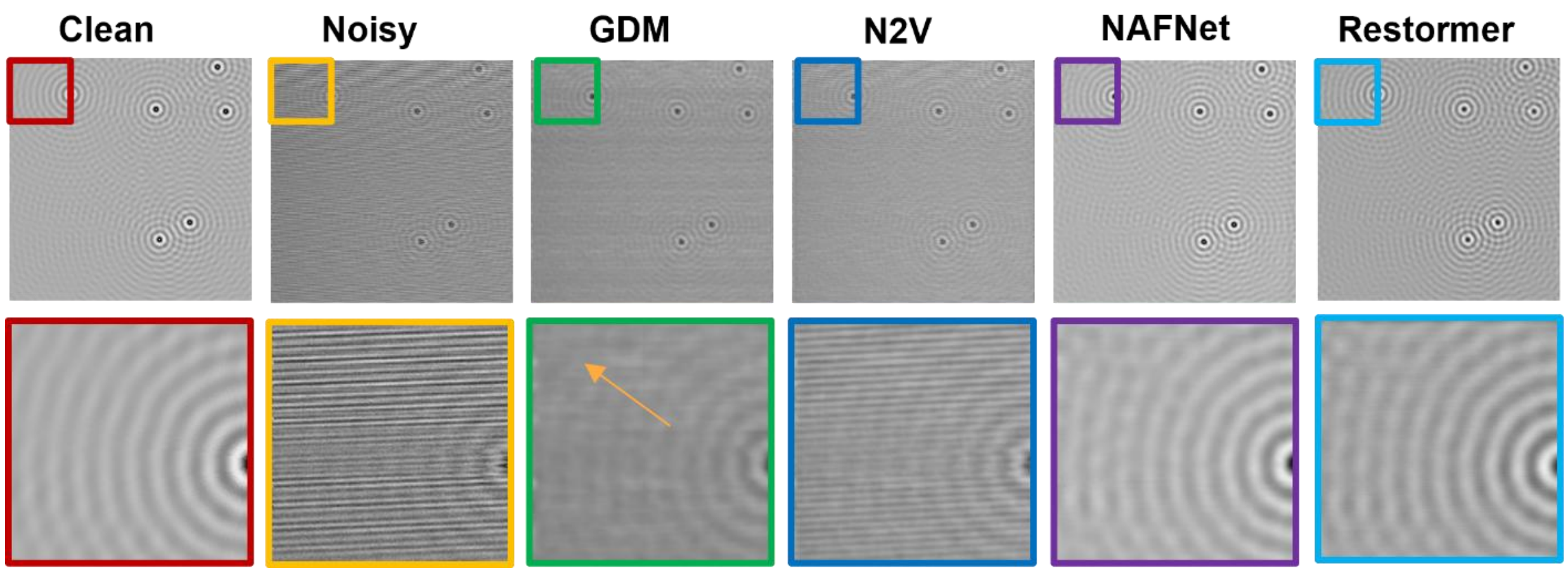

**Figure 3.** Visualization of denoising results on a simulated Cu(111) STM image. From left to right, the simulated clean and respective noisy test image; and the denoised results obtained by unsupervised benchmarked GDM and N2V; and the denoised results obtained by supervised NAFNet and Restormer. Colored boxes in the top row indicate the regions enlarged in the bottom row.

**Table 1** shows the image denoising performance of the stated methods on simulated noisy Cu(111) STM test images where the PSNR and SSIM scores were averaged for 200 test samples. As shown in Table 1, the supervised models achieved higher PSNR and SSIM scores than the unsupervised models, with an average SSIM of $> 0.9$ and PSNR of $> 25$. By contrast, GDM and N2V yielded substantially lower PSNR and SSIM values. This demonstrates an efficient denoising performance by the supervised models on the simulated test set, trained with realistic STM images from our developed toolkit. **Fig. 3** provides a visual representation of the image denoised by different methods, which further supports the stated performance. The enlarged views in the second row, highlighted by the colored boxes, show local details of the denoised results. N2V removes only part of the noise and leaves visible stripe-artifacts. GDM suppresses most of the noise but over-smooths the wave patterns, for example in the upper-left region indicated by the arrow and is therefore still insufficient for detailed analysis of electron wave patterns. In contrast, NAFNet and Restormer produces a denoised image that preserves better details of the wave pattern, as also suggested in Table 1.

### 2.2. Denoising of experimental noisy STM images:

Building on the performance improvement of the supervised ML methods, once trained with simulated realistic STM images generated from the proposed digital framework, we have further evaluated the methods over real experimental STM images. **Fig.4** provides the STM image denoising of Cu(111) from the benchmarked supervised and unsupervised methods, over three different scan sizes obtained under different experimental conditions. It is to be noted that experimental images do not have a ground truth clean image, thus the evaluation has been conducted based on domain-expert visualization and assessments. The colored boxes in the left three columns mark the enlarged regions in the corresponding right panels for each row. By comparing the enlarged patches vertically, we observe that the supervised models consistently outperform the unsupervised models where the former removes the scan artifacts completely while faithfully preserving the underlying physical features such as defects (dark spots) and electron wave patterns (concurrent circles surrounding the defects). This highlights the advantage of supervised training with our developed toolkit to generate large volume of realistic STM data, indicating that the ensemble of noise functions in the simulated images closely resembles that observed in experimental images.

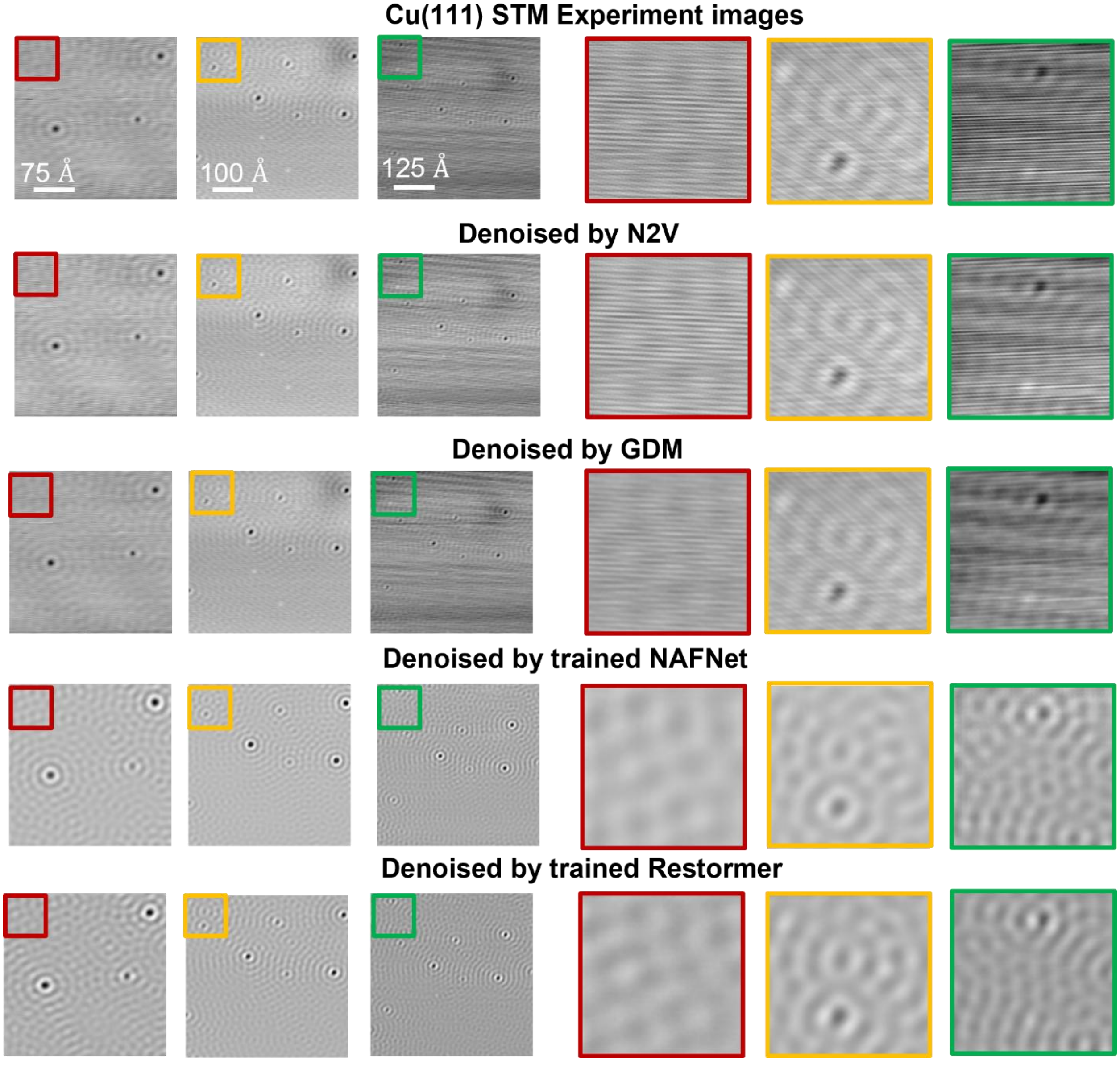

**Figure 4. Denoising comparison on experimental Cu(111) STM images with different scan window sizes.** The top row presents the raw STM images acquired at scan sizes of 300 Å, 400 Å, and 500 Å. The following rows show the results produced by benchmarked unsupervised models such as N2V and GDM; and by benchmarked supervised models such as NAFNet and Restormer trained on our developed digital toolkit generated simulated realistic STM images. For each image, the colored square marks a local region, and the corresponding enlarged patch is displayed on the right using the same color.

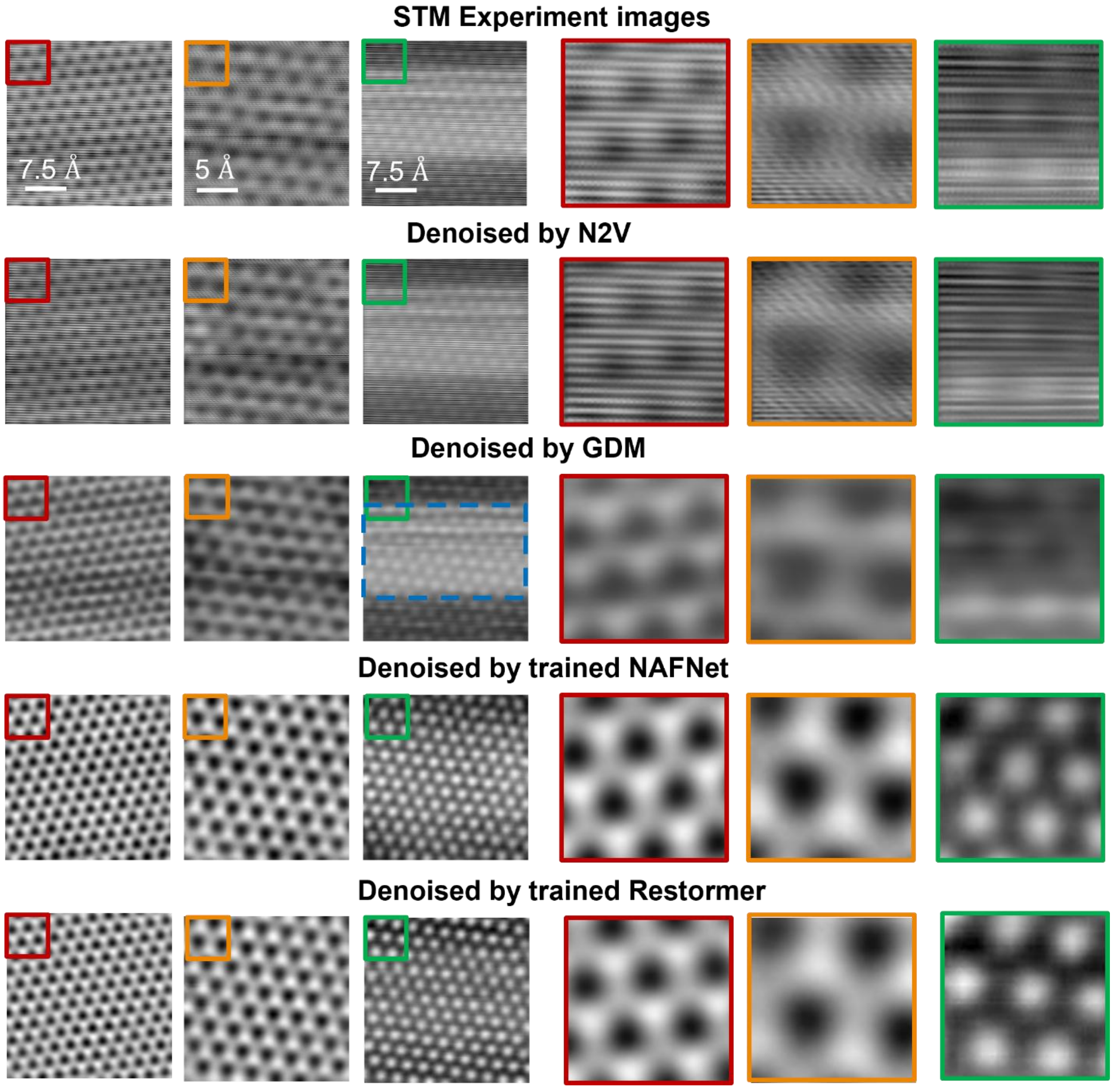

**Figure 5. Denoising comparison on experimental Pb(111) STM images with different scan window sizes.** The top row presents the raw STM images acquired at scan sizes of 30 Å, 20 Å, 30 Å. The following rows show the results produced by N2V, GDM, NAFNet and Restormer trained on simulated STM images. For each image, the colored square marks a local region, and the corresponding enlarged patch is displayed on the right using the same color.

**Fig.5** demonstrates the denoising results of the experimental Pb(111) STM images using benchmarked supervised and unsupervised methods. To add complexity and uncertainty in training images, we considered simulated pair or clean and noisy triangular-lattice images[19]. We considered various contrasts between atomic sites and hollow sites of triangular lattices, because in real microscopy the corruption of artifacts can also change the contrast between these sites. Examples of simulated realistic (noisy) STM images of triangular lattice, generated from our proposed toolkit, have been provided in Supplementary material (**Section S1**). For each image in the left three columns, a colored square marks a selected local region, and the corresponding enlarged patch is displayed on the right in the same color. As per visual inspection, we can clearly see the unsupervised N2V removes little to no noise, while the unsupervised GDM effectively suppresses striped noise but fails to remove the uneven horizontal intensity bands primarily caused by flicker noise. In contrast, the trained supervised NAFNet and Restormer models produces a clean, uniform background with clearly resolved atomic features, while preserving the triangular lattice structures. This indicates that the proposed digitally generated images can adequately represent the noise present in real STM images. In summary, the supervised models trained on the simulated realistic STM images of different material systems outperformed the unsupervised models, expanding the application of supervised methods in microscopy image analysis where experimental data acquisition is expensive and data labelling is infeasible.

### 2.3. Downstream Image Analysis: Learning electron wave patterns of experimental Cu(111) images

From the previous analysis, we have seen that the supervised models trained on rapidly generated noisy-clean STM training images from the proposed digital toolkit can effectively remove noise from experimental STM images and reveal the true underlying physical patterns. In this case study, we demonstrate the impact of such performance in downstream image analysis. **Fig.6** showcased an example of Cu(111) electron wave patterns changing under different magnetic fields. **Fig.6a** shows Cu(111) STM images obtained under 0T ($I_{0T}$), 5T ($I_{5T}$) and 15T ($I_{15T}$) magnetic fields, where **Fig.6b** shows the corresponding images denoised by a trained Restormer model. Figs. 6c-d shows the downstream task analysis of recovering the subtraction between the electron wave patterns for different magnetic fields, before and after supervised denoising. To understand the sensitivity of downstream image analysis performance, we considered two denoising strategies to reveal the subtraction pattern behind the noise. Strategy 1 is subtracting the raw experimental images, then denoise the subtraction; where another strategy 2 is first denoising the experimental images, then subtracting them. We implemented both methods for the pair of experimental images such as $I_{5T}-I_{0T}$ and $I_{15T}-I_{0T}$. To observe the wave shifts induced by the magnetic field, we subtract $I_{5T}-I_{0T}$ and $I_{15T}-I_{0T}$. Without denoising, after subtracting, we can only see artifacts from noise, as shown in the left figure of **Fig.6c** and **Fig.6d**. The middle and right panels of **Figs.6c-d** show the electron wave pattern recognition performance over $I_{5T}-I_{0T}$ and $I_{15T}-I_{0T}$, following strategy 1 and 2 respectively. We can see that the wave pattern looks very similar in the enlarged patches of the same regions as highlighted in the respective middle and right panel images. This consistency confirms that the denoising model reveals true underlying patterns rather than random hallucinations. This example highlights the potential of our digital, realistic STM image simulation toolkit for accelerating scientific discovery. In other words, we have considered the performance of the pattern recovery of the denoised images via removal of existing artifacts and avoiding hallucination of new artifacts. The results open the door for future study on the evolution of electron wave patterns with the magnetic field in various materials to discover new quantum phenomena. It should be noted that, in this paper, we focus on removing noise and revealing the true physical patterns. However, the denoising process altered the scale of images and thus the scale is not exactly comparable between different subtracted images (Fig. 6c and 6d). In future work, we will focus on more accurately preserving the physical scale of the image during denoising.

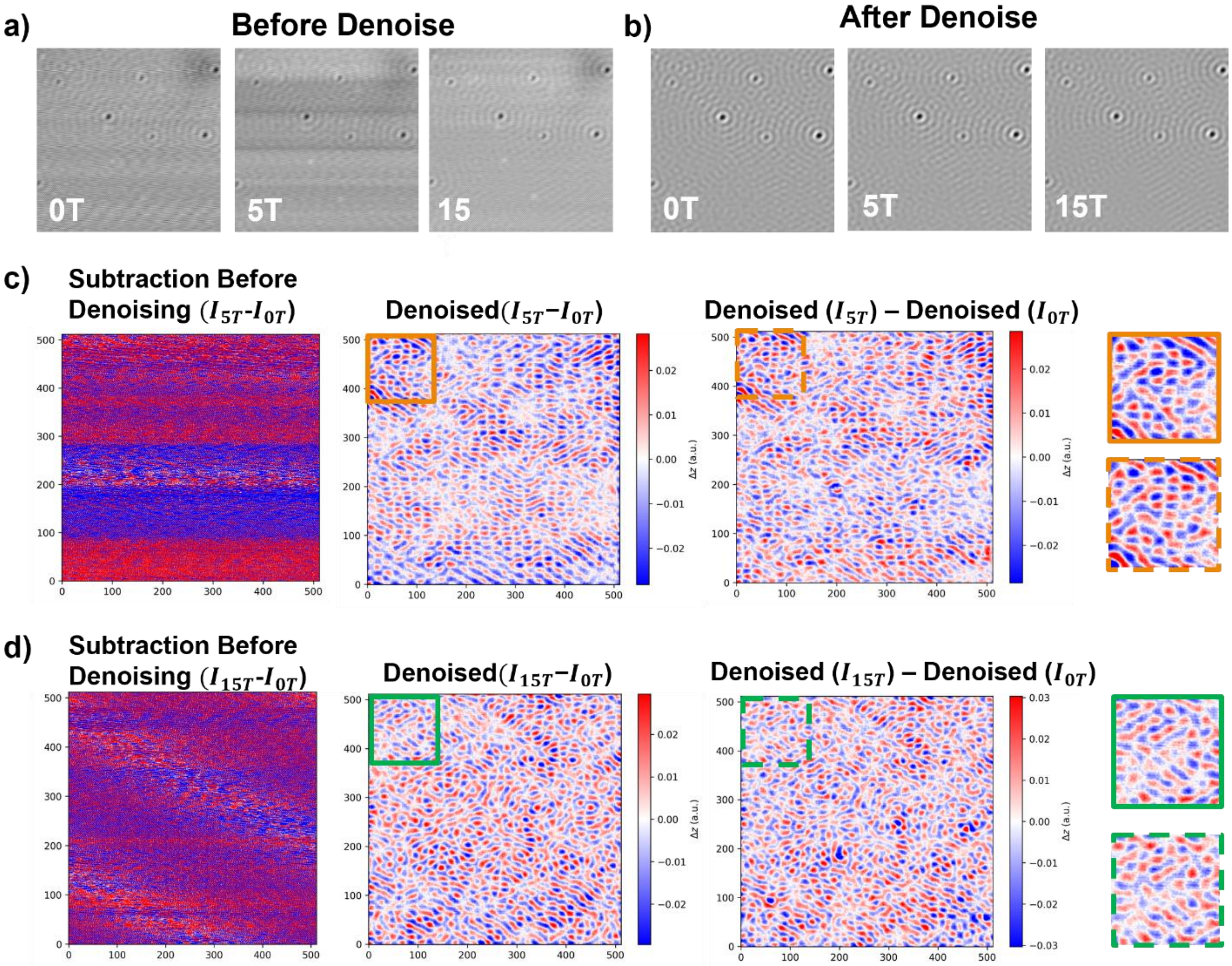

**Figure 6.** Denoising reveals magnetic-field-induced electron wave pattern shifts in Cu(111) STM images. (a) Cu(111) STM images acquired under 0 T ( $\boldsymbol{I_{0T}}$), 5 T ($\boldsymbol{I_{5T}}$), and 15 T ( $\boldsymbol{I_{15T}}$) magnetic fields. (b) Corresponding images denoised by a trained Restormer model. (c) From left to right: subtraction of raw experimental images ($\boldsymbol{I_{5T}}-\boldsymbol{I_{0T}}$); denoised result of the raw subtraction; subtraction of the denoised $\boldsymbol{I_{5T}}$and $\boldsymbol{I_{0T}}$images; and two enlarged patches of the local region marked by the colored squares. (d) From left to right: subtraction of raw experimental images ( $\boldsymbol{I_{15T}} - \boldsymbol{I_{0T}}$ ); denoised result of the raw subtraction; subtraction of the denoised $\boldsymbol{I_{15T}}$ and $\boldsymbol{I_{0T}}$ images; and two enlarged patches of the local region marked by the colored squares.

## 3. Method

### 3.1. Simulation of clean images of Cu(111) and triangular lattices:

Clean Cu(111) and triangular lattice images were simulated using the simulation tool developed by Fiete et al.[20] and Joucken et al.[19] respectively, which we have converted from the original MATLAB package to Python. Each Cu(111) simulated image contains two to eight defects, with a scanning window size of 512 × 512 Å and a resolution of 512 × 512 pixels. Each triangular lattice images have a scanning window size of 30 × 30 Å and a resolution of 256 × 256 pixels. The triangular lattice as simulated using the PyBinding library[21]. The PyBinding library allows us to utilize the single-orbital tight binding model to generate a triangular lattice. This lattice features key components such as scattering centers and a spatial local density of states using the Kernel Polynomial Method (KPM). The lattice is generated using a known lattice constant of 0.35 nanometers and the primitive 2D lattice vectors, $\overrightarrow{a_1} = a(1,0)$ and $\overrightarrow{a_2} = a(\frac{1}{2}, \frac{\sqrt{3}}{2})$. We generate one site per cell and allow for rotation and shift using a 2D rotation and shift matrix. Allowing for different orientations of the triangular lattice to be used as training and test data.

### 3.2. ML Models training procedure

We utilized NAFNet and Restormer as the supervised denoising model. Restormer is a Transformer-based neural network designed for image restoration tasks and has been reported to have top performance in image denoising compared to other image restoration models[22]. NAFNet is another state-of-the-art image restoration model that achieves comparable denoising performance to Restormer while being more computationally efficient. Both models were trained on 1000 pairs of simulated clean-noisy Cu(111) or triangular lattice images, split into training, test, and validation sets (8:1:1). We trained the NAFNet and Restormer models using the AdamW optimizer with a learning rate of $1\times10^{-4}$ and $3\times10^{-4}$ respectively. Training was performed for a maximum of 200 epochs, with early stopping based on validation loss using a patience of 20 and 15 epochs. The batch sizes of the two models differ due to GPU memory limitations. NAFNet was trained with a batch size of 16, whereas the maximum feasible batch size for Restormer was 8. More details are listed in Table 2. During training, the models are saved every 10 epochs, and after the models were fully trained, we selectively chose the right epoch for experimental image denoising by visual examination. The reason is that, during training, the loss function is calculated based on per-pixel value, which is sensitive to the absolute brightness and pixel misalignment whereas in practice we care more about the overall physical pattern structure. Thus, in Fig. 4 and Fig. 5, the NAFNet results were produced using the model saved at the best epoch, whereas the Restormer results were produced using the model saved at epoch 40 in Fig. 4 and Fig. 6 and at the best epoch in Fig. 5.

We utilized N2V and GDM as the unsupervised denoising model. N2V (Noise2Void) is an unsupervised deep learning method developed for microscopy image denoising. It assumes that the noise at a given pixel is statistically independent of the surrounding image content. During training, selected pixels in the noisy image are masked, and the model learns to predict their values based on the surrounding image structure, thereby learning the underlying signal while suppressing unpredictable noise. GDM is another unsupervised denoising model that builds upon the N2V framework and incorporates a Fast Fourier Transform (FFT)-based loss function to provide additional constraints in the frequency domain. GDM and N2V were trained with one enhanced simulation image and one enhanced experimental image, using the default parameter setting from the original papers. The details of models' training parameters used are listed in **Table 2.**

It is important to note that the trained denoising model is sensitive to the physical pixel spacing parameter, which is defined as the actual physical size represented by each pixel. Therefore, the pixel spacing of the input image must match that of the training data. In this study, the Cu(111) training images have a scanning window size of 512 × 512 Å and a resolution of 512 × 512 pixels, corresponding to a physical pixel spacing of 1 Å per pixel. During inference, the noisy image is resampled to 1 Å per pixel to match the training data resolution. The triangular lattice data have a scanning window size of 30 × 30 Å and a resolution of 256 × 256 pixels, corresponding to a physical pixel spacing of 30/256 Å per pixel.

During inference, the noisy image is resampled to 30/256 Å per pixel to match the training data resolution. A detailed sensitivity analysis of model performance with respect to pixel spacing is provided in the supplementary material (**Section S2**).

**Table 2. Hyperparameter details of the applied supervised and unsupervised models during training**

| Model | Patch Size | Batch Size | Epochs | Early stopping | Learning rate | Loss Function | optimizer |
|---|---|---|---|---|---|---|---|
| **GDM** | 128×128 | 8 | 50 | N/A | $1\times10^{-4}$ | MSE+FFT | Adam |
| **N2V** | 64×64 | 8 | 25 | N/A | $4\times10^{-4}$ | MSE | Adam |
| **NAFNet** | 128×128 | 16 | 200 | 20 | $1\times10^{-4}$ | Charbonnier Loss | AdamW |
| **Restormer** | 64×64 | 8 | 200 | 15 | $3\times10^{-4}$ | L1 Loss | AdamW |

**Acknowledgements:**

This work (H.Z) was supported by the University of Tennessee startup funding of A.B. The authors (H.Z and A.B) acknowledge the use of facilities and instrumentation at the UT Knoxville Institute for Advanced Materials and Manufacturing (IAMM) and the Shull Wollan Center (SWC) supported in part by the National Science Foundation Materials Research Science and Engineering Center program through the UT Knoxville Center for Advanced Materials and Manufacturing (DMR-2309083). This work (F.P.) was performed during the 2026 Student Mentoring and Research Training (SMaRT) program provided by The Science Alliance, which is a Tennessee Higher Education Commission (THEC) center of excellence administered by The University of Tennessee-Oak Ridge Innovation Institute (UT-ORII). The STM experiment was supported by the National Science Foundation Materials Research Science and Engineering Center program through the UT Knoxville Center for Advanced Materials and Manufacturing (DMR-2309083) (C.V. and L.B.) and by the University of Tennessee startup funding (W.K.). The authors used a large language model as a coding assistant to draft portions of the simulation/analysis code based on author-provided functional descriptions. All generated code was subsequently tested, refined, and validated by the authors, who take full responsibility for the final software and the results reported in this work.

**Contributions**

H.Z conceived the project while A.B and W.K supervised the project. H.Z designed the digital toolkit, wrote the codes for implementation and analysis, and prepared figures. F.B. assisted H.Z with model training and sensitivity analysis. H.Z, A.B and W.K analyzed and interpreted the results. C.V and L.B prepared experimental STM datasets for model validation. C.V simulated the triangular lattice dataset. W.K supervised and supported funding for C.V and L.B. A.B and H.Z wrote the manuscript while W.K co-wrote the manuscript. All authors have reviewed and provided feedback on the manuscript.

**Conflict of Interest:**

The author confirms there is no conflict of interest

**Code and Data Availability Statement:**

The analysis reported here along with the code is summarized in Notebook for the purpose of tutorial and application to other data and can be found in https://github.com/HuanhuanZhao08/stm_noise

# Supplementary Information

## A Digital Simulation Toolkit for Physics-Based Generation of Realistic Experimental Scanning Tunneling Microscopy Images

Huanhuan Zhao[1*], Laxmi Bhurtel[2,3], Connor Vernachio[2,3], Fahmy Paiziah[4], Wonhee Ko[2,3*], Arpan Biswas[3,5*]

[1]Bredesen Center for Interdisciplinary Research, University of Tennessee, Knoxville, USA, 37996

[2]Department of Physics and Astronomy, University of Tennessee, Knoxville, Tennessee 37996, USA

[3]Center for Advanced Materials and Manufacturing, University of Tennessee, 2641 Osprey Vista Way, Knoxville, Tennessee 37920, USA

[4]Department of Computer Science, Hunter College, City University of New York, 695 Park Avenue, New York, New York 10065, USA

[5]University of Tennessee-Oak Ridge Innovation Institute, University of Tennessee, Knoxville, USA, 37996

**Email**: hzhao31@vols.utk.edu, wko@utk.edu, abiswas5@utk.edu

## S1. Application of the Noise Simulation Toolkit to STM Images of Triangular Lattice

To validate the generalization ability of the proposed noise simulation toolkit, we further tested the denoising pipeline on a triangular lattice dataset. First, 1000 clean triangular lattice images were simulated, each image have a scanning window size of 30 × 30 Å and a resolution of 256 × 256 pixels. Then simulated noise was added using the noise simulation toolkit to generate paired clean-noisy training data. Next, a NAFNet model was trained and applied to experimental triangular lattice STM images for denoising. Figure S1 (a) and (b) are two simulated clean triangular lattice images and their corresponding noisy images.

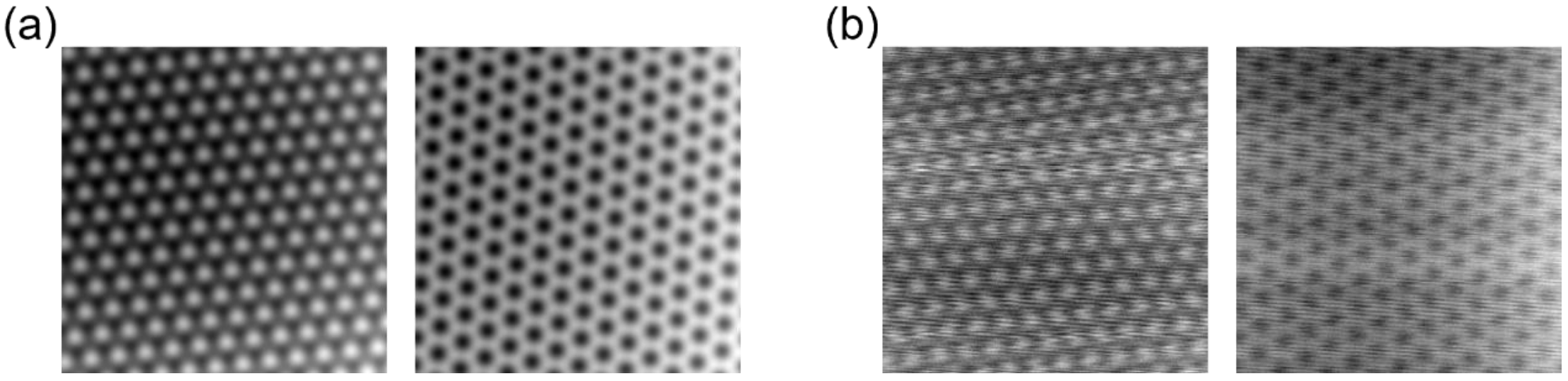

**Figure S1.** Representative examples of simulated clean and noisy STM triangular lattice images

## S2. Sensitive analysis of physical-pixel spacing

When testing the performance of the trained NAFNet model, we found that models trained on simulated images with a fixed scanning window size only perform well when the training and experimental scanning size are matched. Moreover, experimental STM images have scale offset from calibration errors that can be as large as 20 %. To investigate this scale dependence, we trained three separate models using simulated data at scanning window sizes of 270–300 Å, 360–440 Å, and 450–550 Å, respectively, all at the same resolution of 550 × 550 pixels. Each model achieved optimal denoising performance on test images whose scanning window size fell within the corresponding training range, while performance degraded significantly for mismatched scales, as measured by SSIM scores (**Table S1**). Consistently, experimental images acquired at scanning window sizes of 300, 400, and 500 Å were most effectively denoised by the models trained at 270–330 Å, 360–440 Å, and 450–550 Å, respectively (**Fig. S1b, c, d**).

To overcome this scale sensitivity, the noisy input image can be resampled to match the physical pixel spacing of the training data. Specifically, given a model trained on images with a scanning window size of $W_{train}$ Å and resolution of $N \times N$ pixels, the physical pixel spacing is $W_{train}/N$ Å per pixel. For a noisy input image with scanning window size $W_{test}$ Å and the same resolution, the image should be resampled to $W_{test}/(W_{train}/N)$ pixels, so that the physical pixel spacing of the input matches that of the training data. As shown in Table 2, after resampling, model trained with scanning window sizes of 270–300 Å achieved SSIM scores of 0.89 and 0.86 on test images with scanning window sizes of 430 Å and 550 Å, respectively. Model trained with scanning window sizes of 360–440 Å achieved SSIM scores of 0.95 and 0.93 on test images with scanning window sizes of 300 Å and 550 Å, respectively. Model trained with scanning window sizes of 450–550 Å achieved SSIM scores of 0.95 and 0.95 on test images with scanning window sizes of 300 Å and 430 Å, respectively. These results confirm that physical pixel spacing consistently improves model performance. With physical pixel spacing resampling, the model trained with a scanning window size of 450–550 Å maintains an SSIM score above 0.92 across all three test datasets, indicating that a single well-trained model can be used to denoising STM images with any other scanning window sizes. Figure S2(e) shows the denoising results of experimental Cu(111) STM images at different window sizes using the model trained on the 450-550 Å dataset. For the denoising result of Cu(111) presented in main text, we adopted a standardized physical pixel spacing of 1 Å/pixel for all training data and resample experimental images to the same spacing prior to inference. It is important to note that while maintaining a physical pixel spacing close to 1 Å/pixel, the resampled image dimensions must be divisible by 16 due to the NAFNet model encoder-decoder architecture.

**Table S1: Performance of the NAFNet model of experimental Cu(111) images at different scan window sizes.**

| Training window size (Å) | Test: Window size 300 | Test: Window size 430 | Test: Window size 550 |
|---|---|---|---|
| **270 - 300** | **SSIM: 0.91**<br>**PSNR: 26.74** | SSIM: 0.76<br>PSNR: 24.05 | SSIM: 0.72<br>PSNR: 24.75 |
| **360 - 440** | SSIM: 0.78<br>PSNR: 23.72 | **SSIM: 0.90**<br>**PSNR: 25.77** | SSIM: 0.73<br>PSNR: 24.32 |
| **450 - 550** | SSIM: 0.75<br>PSNR: 23.21 | SSIM: 0.82<br>PSNR: 25.73 | **SSIM: 0.92**<br>**PSNR: 28.26** |

**Table S2: Performance of the NAFNet model of experimental Cu(111) images at different scan window sizes, with noisy input image resampling.**

| Training data size (Å) | Test 430 | Test 550 |
|---|---|---|
| **270 - 330** | SSIM: 0.89<br>PSNR: 26.11 | SSIM: 0.86<br>PSNR: 26.56 |
| | **Test 300** | **Test 550** |
| **360 - 440** | SSIM: 0.95<br>PSNR: 26.67 | SSIM: 0.93<br>PSNR: 26.80 |
| | **Test 300** | **Test 430** |
| **450 - 550** | SSIM: 0.95<br>PSNR: 29.89 | SSIM: 0.95<br>PSNR: 28.62 |

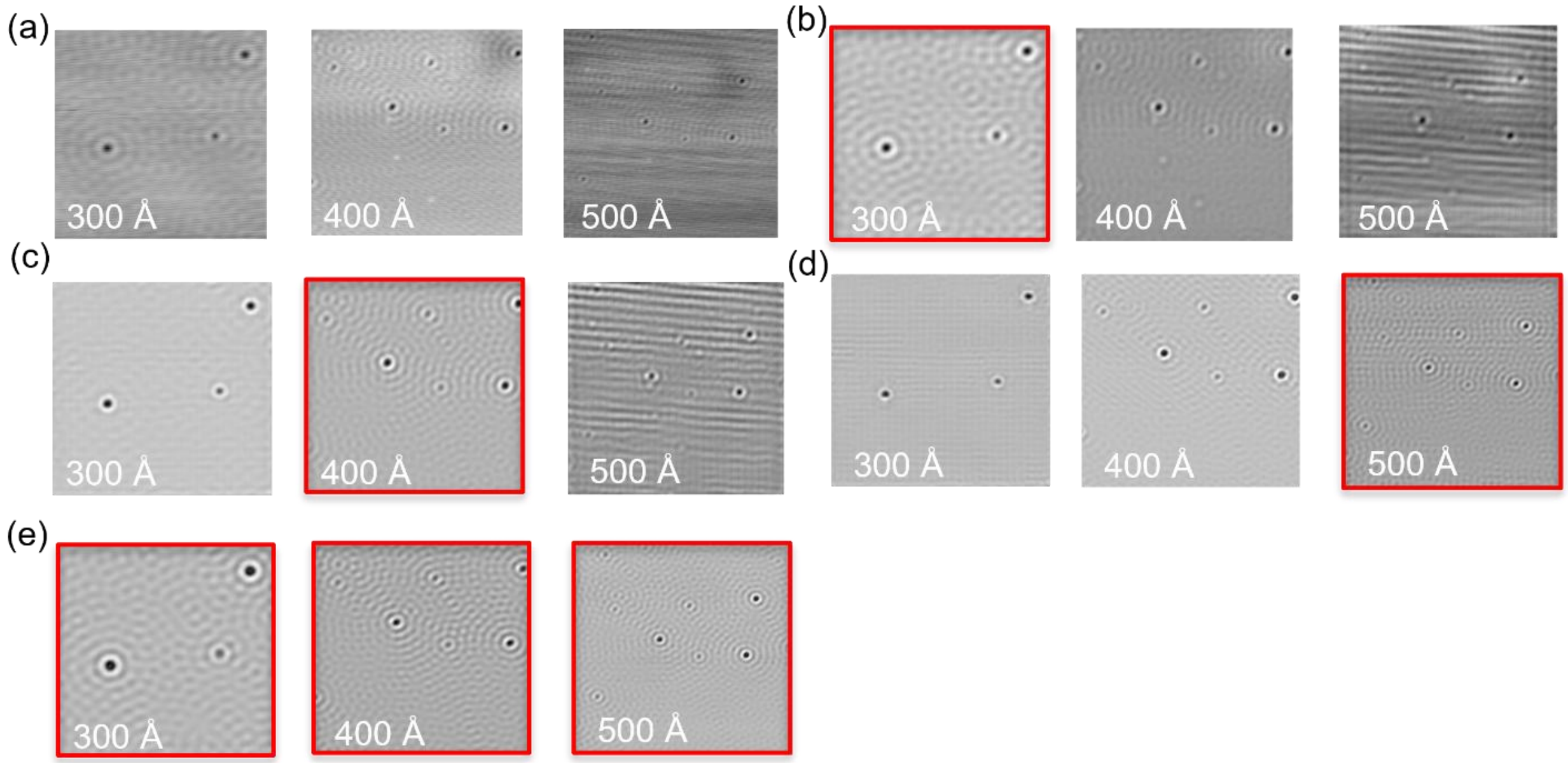

**Figure S2** (a) Three experimental Cu(111) images with different scanning window size, (b) denoising result when training dataset with scanning windows equal [270, 300, 330] Å, no physical pixel spacing resizing. (b) denoising result when training dataset with scanning windows equal [360, 400, 440] Å, no physical pixel spacing resizing. (c) denoising result when training dataset with scanning windows equal [360, 400, 440] Å, no physical pixel spacing resizing. (d) denoising result when training dataset with scanning windows equal [450, 500, 550] Å, no physical pixel spacing resizing. (e) denoising result when training dataset with standard physical pixel spacing 1 Å per pixel, with physical pixel spacing resizing.